\documentclass[sigconf]{acmart}
\usepackage{enumitem}
\usepackage{subcaption}
\usepackage{subcaption}
\AtBeginDocument{%
  }

\copyrightyear{2026}
\acmYear{2026}
\setcopyright{cc}
\setcctype{by}
\acmConference[UbiComp Companion '26]{Companion of the 2026 ACM International Joint Conference on Pervasive and Ubiquitous Computing}{October 11--15, 2026}{Shanghai, China}
\acmBooktitle{Companion of the 2026 ACM International Joint Conference on Pervasive and Ubiquitous Computing (UbiComp Companion '26), October 11--15, 2026, Shanghai, China}
\acmDOI{10.1145/3798063.3837321}
\acmISBN{979-8-4007-2533-3/2026/10}

\usepackage{fancyhdr}   

\fancypagestyle{firstpage}{     
  \fancyhf{}
  \chead{\textcolor{gray}{This article has been accepted for publication in the Companion of the 2026 ACM International \\ Joint Conference on Pervasive and Ubiquitous Computing.}}
  \fancyfoot[C]{\small{\textcolor{gray}{~\copyright~ Personal use of this material is permitted.  Permission from ACM must be obtained for all other uses, in any current or future media, including reprinting/republishing this material for advertising or promotional purposes, creating new collective works, for resale or redistribution to servers or lists, or reuse of any copyrighted component of this work in other works.}}}

}

\begin{document}

\title{EdgeHAR: An Edge-Native Compact Sensor Foundation Model for Human Activity Recognition}


\author{He Zhang}
\affiliation{%
  \institution{Northwestern Polytechnical University}
  \city{Xi'an}
  \state{Shaanxi}
  \country{China}}

\author{Siyu Yuan}
\affiliation{%
  \institution{RPTU University Kaiserslautern-Landau}
  \city{Kaiserslautern}
  \state{Rheinland Pfalz}
  \country{Germany}}

\author{Siyu Liu}
\affiliation{%
  \institution{Xi'an University of Technology}
  \city{Xi'an}
  \state{Shaanxi}
  \country{China}}

\author{Sizhen Bian}
\affiliation{%
  \institution{Northwestern Polytechnical University}
  \city{Xi'an}
  \state{Shaanxi}
  \country{China}}

\author{Bin Guo}
\affiliation{%
  \institution{Northwestern Polytechnical University}
  \city{Xi'an}
  \state{Shaanxi}
  \country{China}}
  
\renewcommand{\shortauthors}{He Zhang, Siyu Yuan, Siyu Liu, Sizhen Bian \& Bin Guo}
\renewcommand{\shortauthors}{He Zhang, Siyu Yuan, Siyu Liu, Sizhen Bian \& Bin Guo}

\begin{abstract}
Sensor-based human activity recognition (HAR) is fundamental to ubiquitous and wearable computing, yet existing foundation models are largely designed for cloud-scale deployment and struggle with real-world sensing shifts, including unseen users, devices, sampling rates, and sensor placements.
We present \textbf{EdgeHAR}, an edge-native compact sensor foundation model designed for wearable intelligence.
Unlike conventional models that entangle activity knowledge with acquisition variations, EdgeHAR learns transferable representations by factorizing sensor signals into three latent codes: an \textbf{(i)Activity-Semantic Code} capturing reusable activity knowledge, a \textbf{(ii)Motion-Dynamics Code} modeling temporal patterns, and an \textbf{(iii)Acquisition-Context Code} representing sensor-specific variations. This disentangled design enables efficient adaptation to new users, devices, placements, and activity classes with limited target-domain data.
By incorporating lightweight adaptation modules, EdgeHAR achieves foundation-model-level transferability while satisfying edge constraints in computation, memory, latency, and privacy. Experiments across heterogeneous HAR datasets demonstrate that EdgeHAR maintains competitive recognition performance under distribution shifts with substantially reduced deployment cost.
EdgeHAR establishes a practical paradigm for compact, edge-first foundation models for ubiquitous sensing systems.
\end{abstract}




\maketitle

\section{Introduction}
\thispagestyle{firstpage} 

Wearable and mobile sensors are fundamental to ubiquitous computing, enabling applications such as health monitoring, rehabilitation, and context-aware interaction~\cite{patsch2025wacu,chen2021deep,rovzanec2023human,zhang2026synthesizing,saha2025feature,ek2025comparing,liang2020behavioral}.
Human activity recognition (HAR) is a key sensing capability, but unlike cloud-based machine learning systems, wearable HAR models must operate directly on resource-constrained edge devices, introducing strict constraints on computation, memory, energy, latency, and privacy.

Recent sensor foundation models have achieved promising transferability through large-scale pretraining and self-supervised learning.~\cite{saha2025pulse,ji2024hargpt,zheng2024heterogeneous,xu2025exploring}
However, existing approaches are primarily designed for cloud-centric deployment, relying on large model capacity, centralized computation, and expensive adaptation.~\cite{bian2026foundation}
Such designs are incompatible with wearable scenarios where models must provide personalized intelligence under limited resources~\cite{narayanswamy2025scaling,qin2023generalizable,bian2022using,bonazzi2024retina,moosmann2024ultra,konwar2025revisiting}.
Furthermore, real-world sensing environments inevitably contain unseen users, devices~\cite{reiss2012introducing,shoaib2014fusion,wang2024optimization}, sampling configurations, sensor placements, and activity categories that are not covered during pretraining.

A key challenge is that sensor signals reflect both human behavior and acquisition conditions~\cite{zhang2026triple}. Variations from device modality, sampling configuration, and body placement are structured factors rather than simple noise.
Existing approaches often treat these variations as domains to align or remove, without explicitly separating reusable activity knowledge from acquisition-specific information.
Consequently, pretrained representations can degrade under deployment shifts, while efficient on-device adaptation remains difficult.

We present \textbf{EdgeHAR}, an edge-native compact sensor foundation model designed for wearable intelligence.
EdgeHAR introduces a structured sensing representation that disentangles activity semantics, motion dynamics, and acquisition context, enabling transferable knowledge reuse across heterogeneous sensing environments.
By combining a compact foundation backbone with lightweight adaptation, EdgeHAR supports efficient personalization on edge devices without full-model retraining or raw-data transmission.

In summary, this work makes three contributions:
\begin{itemize}[topsep=2pt,itemsep=1pt]
    \item We propose \textbf{EdgeHAR}, an edge-native compact sensor foundation model that addresses the gap between foundation-model transferability and wearable deployment constraints.
    \item We introduce a structured disentangled representation that separates activity semantics, motion dynamics, and acquisition context to improve robustness under sensing distribution shifts.
    \item We demonstrate effective transfer across heterogeneous HAR scenarios, including unseen users, devices, placements, sampling configurations, and activity classes, while maintaining favorable accuracy-efficiency trade-offs for edge intelligence.
\end{itemize}

\section{EdgeHAR: An Edge-Native Sensor Foundation Model}

EdgeHAR is an edge-native compact sensor foundation model that learns transferable wearable intelligence by factorizing sensor observations into activity semantics and acquisition-related variations. 
This design enables robust transfer across heterogeneous sensing conditions beyond conventional entangled HAR representations.

\begin{figure*}[t]
    \centering
    \includegraphics[width=\textwidth]{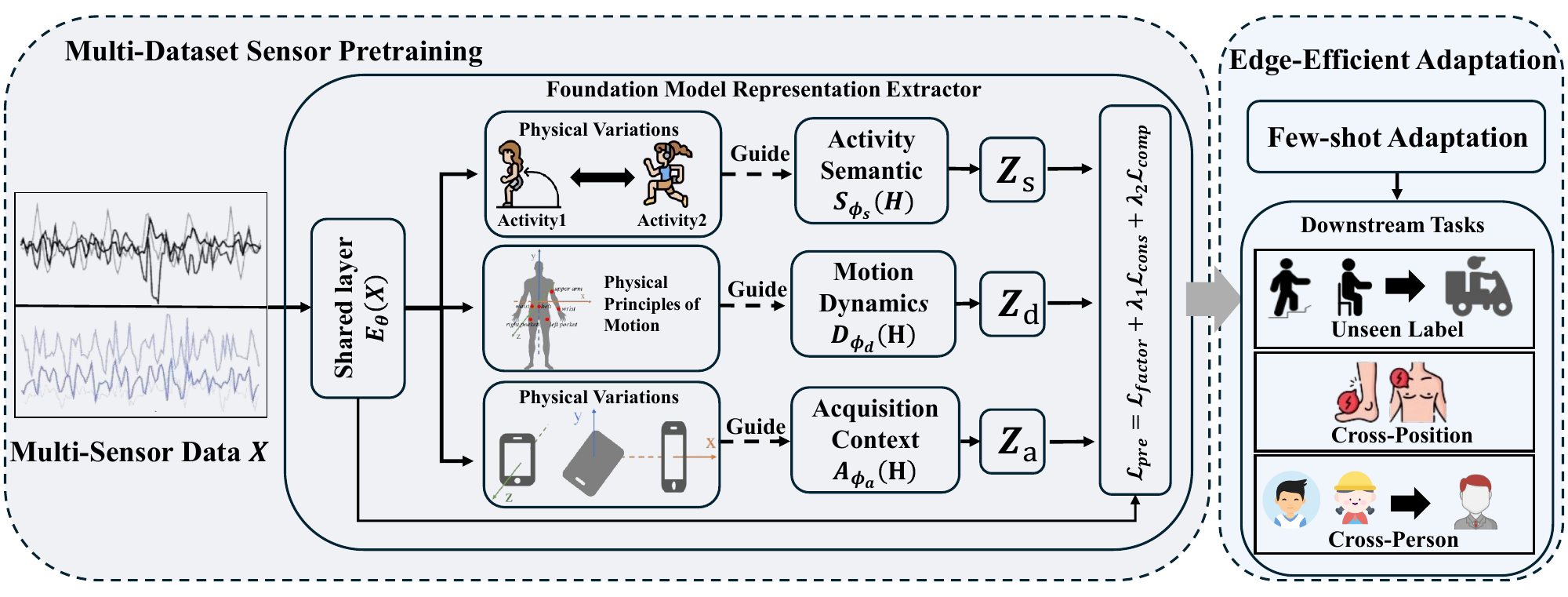}
    \caption{
    Overview of EdgeHAR
    }
    \label{fig:edgehar_framework}
\end{figure*}

As shown in Figure~\ref{fig:edgehar_framework}, the framework consists of two stages: \textbf{(1) Multi-dataset sensor pretraining} for learning generalizable sensing knowledge, and \textbf{(2) Edge-efficient adaptation} for lightweight personalization under deployment constraints.
\subsection{Overview}

Given a sensor window $\mathbf X$, EdgeHAR first extracts a shared representation:

\begin{equation}
\mathbf H=E_{\theta}(\mathbf X).
\end{equation}
Instead of learning a single latent embedding, EdgeHAR decomposes sensor representations into three complementary factors:

    \begin{equation}
(\mathbf z_s,\mathbf z_d,\mathbf z_a)=F(\mathbf H),
\end{equation}
where $\mathbf z_s$, $\mathbf z_d$, and $\mathbf z_a$ represent activity semantics, motion dynamics, and acquisition context, respectively.

The semantic factor captures transferable activity knowledge, while the dynamics and acquisition factors preserve deployment-related variations. This factorization enables robust transfer across unseen sensing conditions.

\subsection{Multi-Dataset Sensor Pretraining}

EdgeHAR is pretrained on heterogeneous wearable datasets containing variations in users, devices, sampling configurations, sensor placements, and activity categories. 
The goal of pretraining is not to remove sensing variations, but to explicitly model their different roles. The semantic factor preserves activity-discriminative information, while dynamics and acquisition factors capture execution and sensing characteristics.

The pretraining objective combines factor-specific learning with representation consistency:

\begin{equation}
\mathcal L_{pre}=\mathcal L_{factor}+\lambda_1\mathcal L_{cons}+\lambda_2\mathcal L_{comp}.
\end{equation}
where $\mathcal L_{cons}$ preserves shared motion information and $\mathcal L_{comp}$ encourages complementary latent factors. This enables EdgeHAR to learn a compact yet transferable foundation representation.

\subsection{Factorized Sensor Representation Learning}

The core design of EdgeHAR is a structured representation that separates three factors underlying wearable sensing:

\textbf{Activity-Semantic Factor.}
The semantic encoder learns activity representations invariant to sensing conditions:

\begin{equation}
\mathbf z_s=S_{\phi_s}(\mathbf H).
\end{equation}

\textbf{Motion-Dynamics Factor.}
The dynamics encoder captures execution-dependent characteristics such as temporal evolution and movement patterns:

\begin{equation}
\mathbf z_d=D_{\phi_d}(\mathbf H).
\end{equation}

\textbf{Acquisition-Context Factor.}
The acquisition encoder models sensing-related variations introduced by device, sampling configuration, and body placement:

\begin{equation}
\mathbf z_a=A_{\phi_a}(\mathbf H).
\end{equation}
By explicitly representing acquisition context, EdgeHAR prevents the semantic representation from absorbing deployment-specific variations. The resulting representation provides a deployment-aware inductive bias for robust sensing transfer.

\subsection{Edge-Efficient Adaptation}

After pretraining, EdgeHAR adapts to target environments using limited labeled data. Instead of full-model fine-tuning, lightweight adapters specialize the semantic representation:

\begin{equation}
\tilde{\mathbf z}_s
=
\mathbf z_s+W_u\sigma(W_d\mathbf z_s).
\end{equation}
Only a small number of parameters are updated during personalization, reducing memory and computation costs.
During deployment, only the compact recognition pathway is retained:

\begin{equation}
f_{edge}(\mathbf X)
=
g(S_{\phi_s}(E_\theta(\mathbf X))).
\end{equation}
This design allows EdgeHAR to transfer knowledge from large-scale heterogeneous pretraining while maintaining an efficient inference architecture suitable for wearable edge devices.
\section{Evaluation: Towards Practical Edge Foundation Intelligence}\
\subsection{Experimental Setup}
We evaluate EdgeHAR under cross-dataset, cross-person, and cross-position shifts. For all settings, only 2\% target-domain samples are used for fine-tuning. Cross-dataset evaluates transfer across unseen datasets, cross-person evaluates unseen users, and cross-position evaluates unseen sensor placements.
\begin{table}[h]
\caption{Statistics of wearable sensing datasets used for EdgeHAR pretraining and evaluation.}
\label{tab:datasets}
\centering
\small
\setlength{\tabcolsep}{4pt}
\begin{tabular}{lccccc}
\toprule
Dataset 
& Subjects 
& Sensors 
& Position 
& Rate
& Classes\\
\midrule
HHAR~\cite{stisen2015smart} 
& 9 
& acc+gyro 
& Multiple 
& 50 HZ 
& 6 \\

MotionSense~\cite{malekzadeh2019mobile} 
& 24 
& acc+gyro 
& Pocket 
& 50 HZ
& 6 \\

UCI-HAR~\cite{reyes2016transition}
& 30 
& acc+gyro 
& Waist 
& 50 HZ 
& 6 \\

Shoaib~\cite{shoaib2014fusion}
& 10 
& acc+gyro+mag 
& Multiple 
& 50 HZ 
& 7 \\

PAMAP2~\cite{reiss2012introducing}
& 9 
& acc+gyro+mag 
& Multiple 
& 100 HZ 
& 18 \\

\bottomrule
\end{tabular}
\end{table}

\subsection{Transferability under Sensing Shifts}

\subsubsection{\textbf{Multi-source cross-dataset}}
We evaluate multi-source transfer by pretraining EdgeHAR on three heterogeneous source datasets and adapting to an unseen target dataset with only 2\% target samples.
As shown in Table~\ref{tab:cross_dataset_multi}, EdgeHAR consistently outperforms existing baselines across different target domains, demonstrating that factorized pretraining enables robust knowledge transfer and efficient personalization under unseen sensing distributions.
\begin{table}[h]
\caption{Multi-source cross-dataset transfer performance (ACC\%). ZS is zero-short and FS is few-short.}
\label{tab:cross_dataset_multi}
\centering
\small
\setlength{\tabcolsep}{3.5pt}
\begin{tabular}{llp{0.7cm}p{1.0cm}p{1.0cm}p{1.2cm}p{1.2cm}}
\toprule
Source & Target 
& SDMix 
& Contrast -Sense 
& CrossHAR 
& \textbf{OUR-ZS} 
& \textbf{OUR-FS} \\
\midrule

S+M+H
& UCI
&75.38&79.43&\underline{88.71}&62.85&\textbf{92.15}\\

U+S+H
& Motion
&64.16&53.20&\underline{76.12}&54.23&\textbf{88.68}\\

H+U+M
& Shoaib
&70.11&63.83&\underline{73.52}&58.62&\textbf{86.28}\\

S+U+M
& HHAR
&71.10&71.75&\underline{74.36}&47.12&\textbf{92.45}\\

\bottomrule
\end{tabular}
\end{table}
\subsubsection{\textbf{Single-source cross-dataset}}
We evaluate single-source transfer by pretraining on one dataset and adapting to unseen target datasets. 
Table~\ref{tab:cross_dataset_single} shows that EdgeHAR consistently outperforms existing Transformer and sensor foundation baselines across diverse source-target pairs.
With only 2\% target-domain samples for adaptation, EdgeHAR achieves substantial gains over zero-shot transfer, demonstrating strong representation transferability and efficient personalization under unseen sensing distributions.
\begin{table}[h]
\caption{Single-source cross-dataset transfer performance (ACC\%). ZS is zero-short and FS is few-short.}
\label{tab:cross_dataset_single}
\centering
\small
\setlength{\tabcolsep}{3.5pt}
\begin{tabular}{llp{1.2cm}p{0.8cm}p{1.0cm}p{1.0cm}p{1.1cm}}
\toprule
Source & Target 
& Transformer 
& UniHAR 
& LLM4HAR 
& OUR-ZS 
& OUR-FS \\
\midrule

UCI
& Shoaib 
&25.99&69.90&\underline{74.98}&63.32&\textbf{90.70}\\
& Motion 
&35.64&76.08&\underline{87.09}&62.45&\textbf{88.23}\\
& HHAR 
&44.52&75.43&\underline{85.75}&61.08&\textbf{96.08}\\

\midrule

Motion
& UCI
&39.36&69.05&\underline{89.44}&67.96&\textbf{93.09}\\
& Shoaib
&44.26&78.71&73.70&\underline{77.64}&\textbf{93.01}\\
& HHAR
&42.97&58.33&75.21&\underline{79.68}&\textbf{97.53}\\

\bottomrule
\end{tabular}
\end{table}
\subsubsection{\textbf{Cross-person Transfer}}
To evaluate user generalization, we train EdgeHAR on one group of users and test on unseen users across multiple datasets. Table~\ref{tab:cross_person} shows that EdgeHAR-FS consistently outperforms existing transfer approaches, achieving the highest average accuracy across datasets. The improvement from EdgeHAR-ZS to EdgeHAR-FS with only 2\% target data highlights the effectiveness of EdgeHAR's transferable representation and efficient user personalization.
\begin{table}[t]
\caption{Cross-person transfer performance (ACC\%) on unseen users.}
\label{tab:cross_person}
\centering
\small
\setlength{\tabcolsep}{4pt}
\begin{tabular}{lcccc}
\toprule
Method 
& UCI 
& Motion 
& HHAR 
& Avg. \\
\midrule

SDMix
& 90.33
& 75.48
& 77.44
& 81.08 \\

ContrastSense
& \textbf{93.33} 
& 83.13
& 87.66
& 88.04\\

DI2SDiff
& 87.68
& \underline{88.43}
& \underline{89.19}
& \underline{88.43} \\

\textbf{OUR-ZS}
& 86.19& 85.42& 88.97& 86.86 \\

\textbf{OUR-FS}& \underline{91.88}& \textbf{93.17}& \textbf{94.89}& \textbf{93.31} \\

\bottomrule
\end{tabular}
\end{table}
\subsubsection{\textbf{Cross-position Transfer}}
Sensor placement is a major source of acquisition shift in wearable sensing.
We evaluate this challenge using leave-one-position-out transfer on the Shoaib dataset, where one position is held out as the target domain and only 2\% target samples are provided for adaptation. Table~\ref{tab:cross_position} shows that EdgeHAR consistently outperforms existing transfer methods across unseen positions.
The strong zero-shot performance indicates robust acquisition-aware representation learning, while further gains after few-shot adaptation demonstrate efficient personalization under limited edge data.
\begin{table}[h]
\caption{Cross-position transfer performance (ACC\%) on shoaib dataset. ZS is zero-short and FS is few-short.}
\label{tab:cross_position}
\centering
\small
\setlength{\tabcolsep}{3.5pt}
\begin{tabular}{llccccc}
\toprule
Source & Target 
& Mixup 
& ContrastSense 
& DI2SDiff 
& \textbf{OUR-ZS} 
& \textbf{OUR-FS} \\
\midrule

B+P+W
& Arm
&40.17&70.81&\underline{74.14}&6.00&\textbf{95.71}\\

A+P+W
& Belt
&39.87&68.06&\underline{68.71}&19.90&\textbf{88.29}\\

A+B+W
& Pocket
&47.01&71.48&\underline{72.36}&69.25&\textbf{90.19}\\

A+B+P
& Wrist
&41.12&75.54&\underline{77.11}&42.09&\textbf{91.59}\\

\bottomrule
\end{tabular}
\end{table}
\subsection{Edge Efficiency}
EdgeHAR is designed to achieve foundation-model transferability under strict edge constraints. 
Table~\ref{tab:efficiency} compares deployment costs in terms of model size, parameters, memory, and inference latency.
Large-scale foundation models introduce substantial overhead, while compact HAR models often lack transferable representation capability.
\begin{table}[h]
\caption{Efficiency comparison under edge deployment.}
\label{tab:efficiency}
\centering
\small
\setlength{\tabcolsep}{5pt}
\begin{tabular}{lcccc}
\toprule
Model 
& Params(M)
& Size(MB)
& CPU(ms)
& Memory(MB)\\
\midrule

GPT-2 Small
&160.0&400.0&1028.0&--\\
LLMHAR
&8.0&46&104.5&--\\

\midrule

CNN
&6.45&25.805&2.14&51.05\\
DeepConvLSTM
&0.46&1.85&40.90&10.16\\
HART
&1.45&5.91&5.38&12.74\\
MobileHART
&2.54&10.35&13.49&20.77\\

\midrule

\textbf{EdgeHAR (Full)}
&1.41&5.41&89.8&21.37\\
\textbf{EdgeHAR (Deploy)}
&1.086&4.14&79.0&18.42\\

\bottomrule
\end{tabular}
\end{table}
EdgeHAR achieves a compact deployment footprint with 1.41M parameters and 5.41 MB storage in the full model.
The deployment version further reduces the size to 1.086M parameters and 4.14 MB while requiring only 18.42 MB memory. 
Lightweight adaptation enables efficient personalization without full-model updating, demonstrating a favorable accuracy-efficiency trade-off for wearable edge intelligence.
\subsection{Adaptation under Limited Target Data}

\subsubsection{\textbf{Few-shot Emerging Activity Adaptation}}
We evaluate label-shift adaptation by training EdgeHAR on known activities and adapting to unseen PAMAP2 activities. With only 100 samples per emerging activity, EdgeHAR achieves 91.86\% Macro-F1, demonstrating that the learned semantic representation supports efficient few-shot adaptation beyond the pretraining label space.
\begin{table}[h]
\caption{Few-shot adaptation to emerging activities on PAMAP2. 
K denotes the number of target samples per activity.}
\label{tab:emerging_activity}
\centering
\small
\setlength{\tabcolsep}{6pt}
\begin{tabular}{lccc}
\toprule
Emerging Activity
& K=20
& K=50
& K=100\\
\midrule

Lying back
&0.0022
&0.6674
&0.8611\\

Walking parking lot
&0.3260
&0.8871
&0.9899\\

Cycling bike horizontal
&0.0596
&0.3849
&0.9413\\

Playing basketball
&0.2471
&0.6147
&0.8786\\

\midrule
Macro-F1
&0.1627
&0.7487
&0.9186\\

Overall Accuracy
&0.2382
&0.7574
&0.9190\\

\bottomrule
\end{tabular}
\end{table}

\subsubsection{\textbf{Few-shot Adaptation Efficiency}}
\begin{figure}[t]
    \centering
    \begin{subfigure}{0.49\columnwidth}
        \centering
        \includegraphics[width=\linewidth]{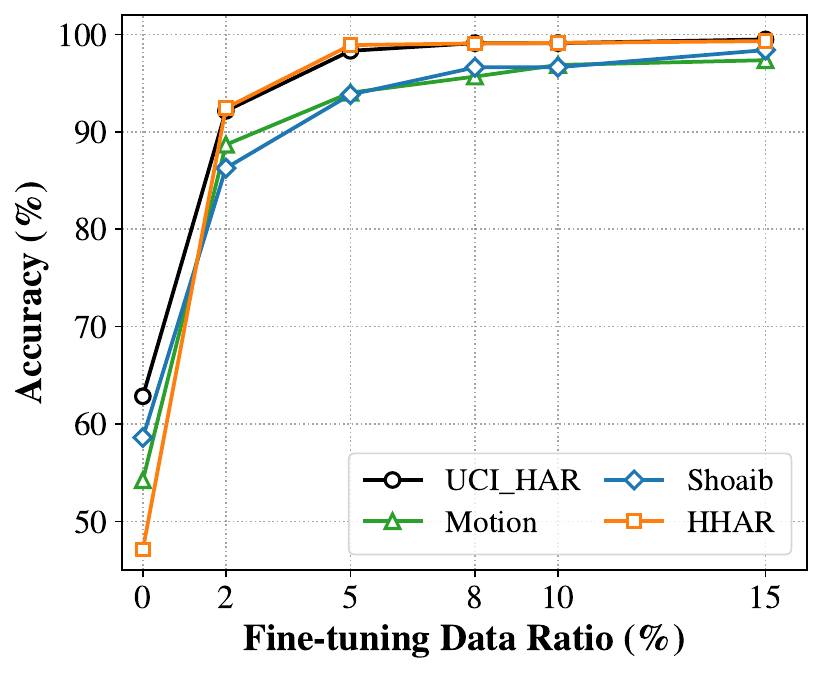}
        \caption{Accuracy}
        \label{fig:finetune_acc}
    \end{subfigure}
    \hfill
    \begin{subfigure}{0.49\columnwidth}
        \centering
        \includegraphics[width=\linewidth]{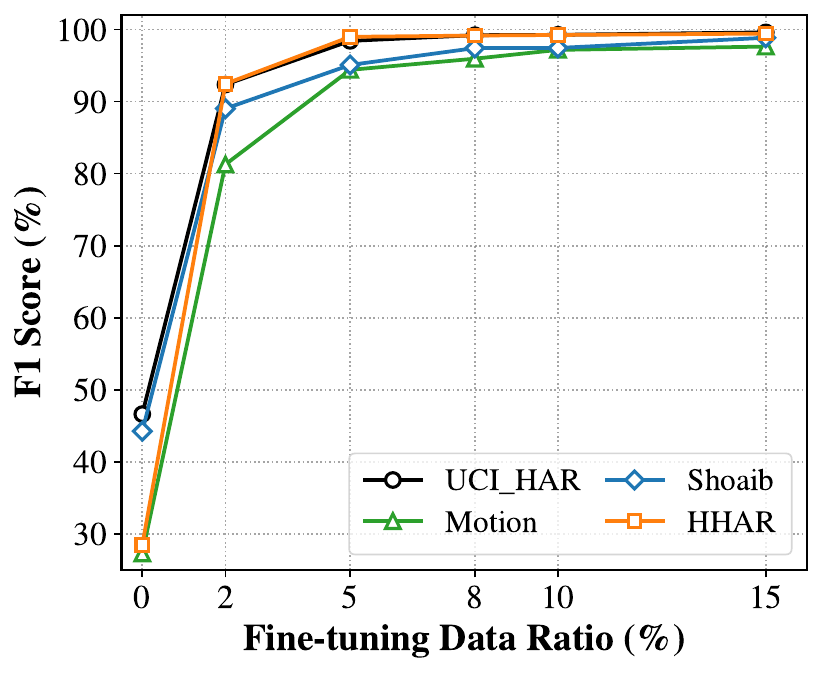}
        \caption{F1-score}
        \label{fig:finetune_f1}
    \end{subfigure}
    \caption{Effect of target-domain fine-tuning data ratio on adaptation performance. 
    EdgeHAR is evaluated under different amounts of target data across heterogeneous datasets.}
    \label{fig:data_efficiency}
\end{figure}

We investigate how target data availability affects EdgeHAR personalization. By varying the fine-tuning ratio from 0\% to 15\%, we evaluate adaptation under different annotation budgets. As shown in Figure~\ref{fig:data_efficiency}, EdgeHAR rapidly improves with increasing target-domain data ratios across different datasets. Even with limited fine-tuning samples, the model achieves substantial performance gains, demonstrating efficient adaptation under constrained annotation budgets.
The strong performance under extremely limited supervision demonstrates that EdgeHAR can effectively reuse pretrained sensing knowledge and achieve efficient on-device adaptation.


\section{Discussion and Conclusion}
This work presents EdgeHAR, an edge-native compact sensor foundation model for wearable HAR. By disentangling activity semantics and acquisition variations with lightweight adaptation, EdgeHAR enables transferable edge intelligence. Extensive experiments across unseen users, devices, placements, and sampling rates confirm its robustness, while its small footprint supports real‑time on‑device inference. Our study shows that structured representation learning can provide an effective path toward efficient and personalized ubiquitous foundation models.

\begin{acks}
This work is supported by the Fundamental Research Funds for the Central Universities.
\end{acks}

\balance
\bibliographystyle{ACM-Reference-Format}
\bibliography{sample-base}

\end{document}